\documentclass[runningheads]{llncs}
\usepackage[T1]{fontenc}

\usepackage{graphicx}
\usepackage{multirow}
\usepackage{amsmath}
\usepackage{esvect}

\begin{document}
\title{Comparing the Performance of LLMs in RAG-based Question-Answering: A Case Study in Computer Science Literature}
\titlerunning{Performance of LLMs in RAG-based Question-Answering}

\author{Ranul Dayarathne\inst{1}\orcidID{0009-0008-8541-6086} \and
Uvini Ranaweera\inst{1}\orcidID{0009-0003-6205-3482} \and
Upeksha Ganegoda\inst{1}\orcidID{0000-0002-1745-9112}}
\authorrunning{R. Dayarathne et al.}
\institute{University of Moratuwa, Katubedda, Moratuwa 10440, Sri Lanka
\email{\{dayarathnedvrn.19,ranaweeraraua.19,upekshag\}@uom.lk}}
\maketitle  

This version has been accepted for publication after peer review, but is not the Version of Record. The Version of Record is available at https://doi.org/10.1007/978-981-97-9255-9\_26. Use of this Accepted Version is subject to Springer’s terms of use.
\begin{abstract}
Retrieval Augmented Generation (RAG) is emerging as a powerful technique to enhance the capabilities of Generative AI models by reducing hallucination. Thus, the increasing prominence of RAG alongside Large Language Models (LLMs) has sparked interest in comparing the performance of different LLMs in question-answering (QA) in diverse domains. This study compares the performance of four open-source LLMs, Mistral-7b-instruct, LLaMa2-7b-chat, Falcon-7b-instruct and Orca-mini-v3-7b, and OpenAI’s trending GPT-3.5 over QA tasks within the computer science literature leveraging RAG support. Evaluation metrics employed in the study include accuracy and precision for binary questions and ranking by a human expert, ranking by Google's AI model Gemini, alongside cosine similarity for long-answer questions. GPT-3.5, when paired with RAG, effectively answers binary and long-answer questions, reaffirming its status as an advanced LLM. Regarding open-source LLMs, Mistral AI’s Mistral-7b-instrucut paired with RAG surpasses the rest in answering both binary and long-answer questions. However, among the open-source LLMs, Orca-mini-v3-7b reports the shortest average latency in generating responses, whereas LLaMa2-7b-chat by Meta reports the highest average latency. This research underscores the fact that open-source LLMs, too, can go hand in hand with proprietary models like GPT 3.5 with better infrastructure.

\end{abstract}

%INTRODUCTION

\section{Introduction}

Looking back, if one wanted to query something a few years back, the go-to solution would be “Google”, and would end up reading through many retrieved web sources to find the solution. Nowadays, Large Language Models (LLMs) like GPT have made our lives easy, and is a common thing to say, “Let me ask Chat-GPT for the answer”. Thus, by now, every one of us has tackled different applications built on top of Generative Artificial Intelligence (GenAI)~\cite{a16} and has witnessed how they do a fair job in providing us with the required information.

Since almost all the  LLMs released so far are trained on billions of parameters on top of massive amounts of data, they can perform most of the natural language tasks and cover an abundance of knowledge sources that were available back when the models were trained. Indirectly, this implies that if an LLM is not regularly trained, it cannot retrieve the most up-to-date theories in the field. For example, the latest update of ChatGPT 3.5 includes knowledge sources only up to January 2022~\cite{a17}, making it incapable of correctly capturing theories and concepts published after that. This has brought forward the problem of hallucination, where a model generates confident answers that are factually incorrect~\cite{a18}. Therefore, it is not always a better option to use vanilla LLMs to query for unknown facts, and most importantly, a vanilla LLM won’t be an ideal option to query for unique concerns as they are not customised to fit such. 

In order to customise LLMs to fit user needs, Retrieval Augmented Generation (RAG) is used widely around the globe. Patrick Lewis~\cite{a19} introduced RAG for Natural Language Processing (NLP) tasks in 2021, proposing to combine non-parametric memory with pre-trained parametric-memory generation models. In simple terms, RAG allows an LLM to retrieve data from external databases specified by the user and keep up with the latest findings in the field. This has brought many opportunities for multiple domains, and only a handful of studies are being conducted to explore the applicability of RAG to enhance domain-specific knowledge extraction~\cite{a14,a4,a6}. 

One prominent field that can benefit from RAG-based LLM applications is education; to be precise, RAG can be introduced to enhance the accessibility of scholarly sources. Scholarly sources are works published by experts in a field to share new findings and theories~\cite{a20}. The rapid pace at which scholarly sources evolve is beyond an individual’s time capacity to read through and comprehend the findings. Thus, incorporating RAG on top of an LLM makes it possible to facilitate Question-Answering (QA) so that the researchers can keep up with the domain. 

When it comes to research fields, they spread into numerous branches, where computer science is one of the fields with the most significant scholarly contributions~\cite{a21}. Contributions in the computer science domain span from foundational theories and algorithms to groundbreaking technologies such as Artificial Intelligence (AI), which has revolutionised nearly every aspect of modern life. In recent years, the computer science field has experienced a bloom owing to notable advancements in AI, Quantum Computing, Robotics, Bioinformatics, Virtual Reality, Augmented Reality, etc~\cite{a22}. This sudden bloom has opened the door for scientists worldwide to explore even beyond and contribute to further advancements, leading to a significant increase in scholarly sources. Thus, customising a language model to perform RAG on scholarly sources on computer science is no disadvantage, as it will help more people easily identify new trends in the field and update their knowledge. 

This paper presents a RAG-enabled QA system for the latest inventions in the computer science domain. While RAG methodologies have found widespread use, the integration and performance evaluation in the computer science domain remains uncharted. The study fills the gap by conducting a comparative analysis of the performance of multiple LLMs, including  GPT 3.5, LLaMa-2-7b-chat, Mistral-7b-instruct, Falcon-7b-instruct and Orca-mini-v3-7b, with and without RAG integration. That’s not all; apart from answer quality, it is vital to understand the usefulness, cost and efficiency of an answer. Thus, the study expands its scope to cover those aspects of an LLM while employing a meticulously constructed database sourced from the latest publications in the computer science domain.

The rest of the research paper is structured as follows. Section~\ref{sec:Literature Review} reviews the literature on RAG-based applications. Section~\ref{sec:Methodology} includes the methodology, which presents the conceptual framework along with an overview of the dataset, pre-processing steps, an overview of vectorising the dataset, steps to build the QA pipeline and the performance measures used to evaluate the LLMs. Section~\ref{sec:Discussion} carries out an extensive comparison of the results obtained using different LLMs, followed by section~\ref{sec:Conclusion}, which discusses the limitations of the study and suggestions to enhance the work. 

%LIT REVIEW

\section{Literature Review}
\label{sec:Literature Review}
The applications of LLMs span from search engines to customer support to translation and encompass a broad array of fields~\cite{a1}, as the capacity of an LLM is immense when generating contextually fitting responses~\cite{a2}. Ever since LLMs became popular, RAG received attention from scholars as a remedy to mitigate the hallucination that comes with an LLM due to its inability to grab the most up-to-date data~\cite{a3}. 

Y. Hicke et al.~\cite{a4} introduced AI-TA, which is an intelligent QA assistance for online QA platforms. They have evaluated the performance of their QA pipeline incorporating LLaMa family’s LLaMA-2-13B (L-13) model, LLaMA-2-70B model and OpenAI’s GPT-4 with augmenting techniques such as RAG, supervised fine-tuning (SFT), and Direct Preference Optimization (DPO). The results showed that the combination of L-13 with SFT, DPO and RAG showed the most promising results, with a 107\% increment in accuracy and a 177\% increment in usefulness against the base model performance.

As RAG remains one of the latest advancements in language modelling, there is little research concerning enhancing domain-specific knowledge of LLMs with RAG, where most of the existing research is focused on the biomedical field. Quidwai et al.~\cite{a5} proposed a pipeline that incorporates RAG on top of the Mistral-7B model for QA on multiple myeloma. Markey et al.~\cite{a6} carried out a study to show how clinical trial documentation can be improved with RAG. L\'ala et al.~\cite{a7} proposed PaperQA, which utilised a RAG incorporated LLM to answer questions on biomedical literature. The evaluation of PaperQA’s performance on the custom questions showed that it outperformed not only the LLMs such as Claude-2 and GPT-4 but also the commercial tools Elicit, Scite, Perplexity and Perplexity (Co-pilot), which are specifically designed to deal with scientific literature. There are several more studies focused on specific areas of medicine, such as nephrology~\cite{a9}, liver diseases~\cite{a10}, chest-x rays~\cite{a11}, ECG diagnosis~\cite{a12} etc. 

Moving to other domains, Wiratunga et al.~\cite{a13} introduced CBR-RAG, which utilises case-based reasoning for legal QA, while Chouhan et al.~\cite{a14} presented LexDrafter, which assists in drafting definition articles using RAG. Zhang et al.~\cite{a15} presented a framework to predict the sentiment of financial incidents by incorporating RAG and instruction-tuned LLMs, where the results improved the accuracy of off-the-shelf LLMs like ChatGPT and LLaMa by 15\% to 48\%. 

To date, there is no/less evidence of research proposing RAG to enhance scholarly knowledge accessibility of the computer science domain; and to be more precise, no evidence is present in the computer science domain that evaluates the performance of RAG-incorporated open-source LLMs and closed-source LLMs.

\section{Methodology}
\label{sec:Methodology}

\begin{figure}[ht]
\includegraphics[width=\textwidth]{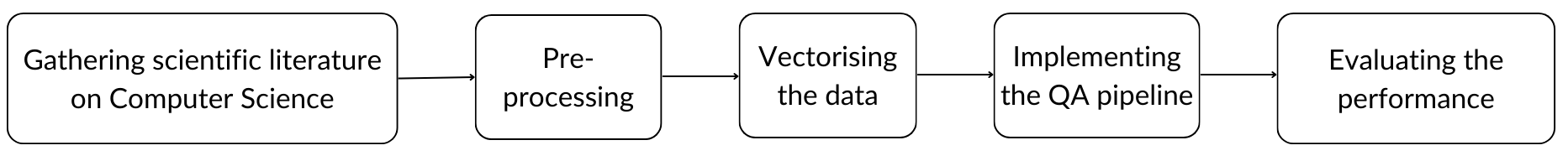}
\caption{Conceptual framework}\label{Fig-1}
\end{figure}

As indicated in Fig.~\ref{Fig-1}, a conceptual framework was designed to address the research aim of evaluating the performance of LLMs in RAG-based QA in computer science literature. 

\subsection{Preparing the Dataset}
Since off-the-shelf LLMs cannot perform RAG when answering user queries and hence need to be customised to perform RAG, much attention was paid to including the latest developments in the field when preparing the dataset. This will help us demonstrate the importance of RAG over off-the-shelf LLMs that are designed as parametric memory generation models~\cite{a19}.

Due to the limitations in accessing full papers and the limited computational resources, it was decided to include only the abstracts of computer science journal papers. This decision relies on the assumption that an abstract of a paper summarises the important findings of an entire study~\cite{a27}. Moreover, Springer and IEEE were chosen as sources to extract abstracts owing to their consistent reputation for publishing high-quality work in technology~\cite{a23}. The compiled dataset had 4929 abstracts from computer science journal papers focusing on large language models, quantum computing and edge computing as they are the latest trends in the domain~\cite{a24}. The abstracts were extracted only from papers published between 2023 - 2024 to ensure that they are the latest discussions and that LLMs are not trained on top of those discussions. The dataset included 1121 abstracts for large language models, 1810 abstracts for edge computing and 1998 abstracts for quantum computing.

% \begin{figure}[ht]
% \includegraphics[width=\textwidth]{Images/fig3.png}
% \caption{Wordclouds for research areas} \label{Fig-3}
% \end{figure}

Along with the abstracts, it was decided to extract the title, author/s, publishing date and as well as keywords to support any future requirements such as retrieval based on metadata. 

\subsection{Pre-Processing}

Immediately after compiling the dataset, a few pre-processing steps were applied to refine the data further. Some excerpts contained irrelevant HTML characters since the abstracts were extracted from the World Wide Web (WWW). Thus, during pre-processing, it was ensured to get rid of such noise in the dataset.
 
Apart from the major fix mentioned above, general pre-processing steps like lower casing, dropping duplicated records, removing null values and removing irrelevant characters were carried out to ensure that the dataset is standardised with no/little noise. To detect and eliminate irrelevant characters such as colons, semi-colons and extra white spaces, along with HTML tags, the Python library “Regular Expressions” (re) was used.   

The application of the above pre-processing steps introduced a standard format to abstracts with zero noise from unnecessary characters. Since the dataset will be used on top of GenAI, no further pre-processing steps were applied, as the LLMs already have the capability to grab the context. Following the pre-processing steps, all the abstracts were stored as individual files in JSON format. 

\subsection{Vectorising the Data}

To perform RAG using LLMs, it is a must to let LLM access an external data source (a non-parametric memory)~\cite{a19}. In our case, the external data source will be the collection of abstracts from computer science journal articles. 

\begin{figure}[ht]
\includegraphics[width=\textwidth]{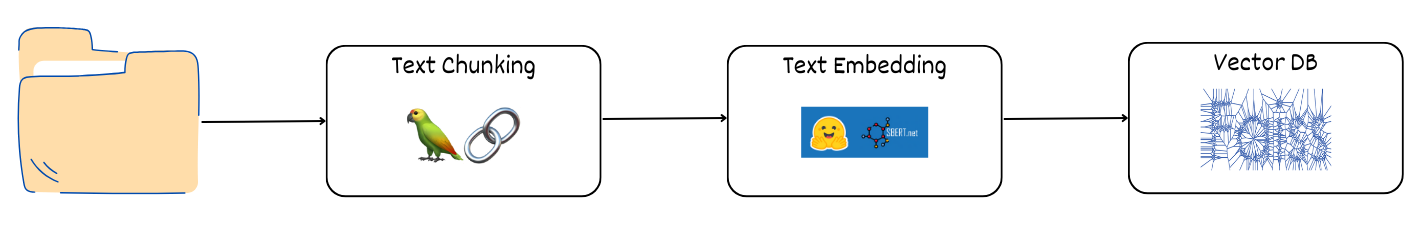}
\caption{Conversion process of the abstracts} \label{Fig-4}
\end{figure}

Having these text files as they are in a database won’t smooth the retrieval process. Thus, converting them into a format that language models can process was necessary~\cite{a41}. This was the point where text embedding came into the spotlight. As defined by Rajvardhan Patil et al.~\cite{a25}, text embedding involves converting textual data into a numerical format (precisely vectors) that machines can process. Throughout the entire process of vectorising the data (Fig.~\ref{Fig-4}), different features offered by the LangChain framework were used~\cite{a26}. 

The conversion process included splitting standardised text excerpts into chunks of 1024 characters maximum. The chunking limit was set such that the content wouldn’t exceed the context window of any LLM considered in the underlying study. To minimise the harm to the context of the abstract during the split, double line breaks ("\textbackslash{n}\textbackslash{n}") and full stops followed by a space (". ") were defined as splitting points. Moreover, a chunk overlap of 200 characters was allowed to ensure continuity among chunks. All in all, LangChain’s "RecursiveCharacterTextSplitter"~\cite{a26} was utilised for chunking as it allows content splits based on a custom set of characters. 

\begin{figure}[ht]
\includegraphics[width=\textwidth]{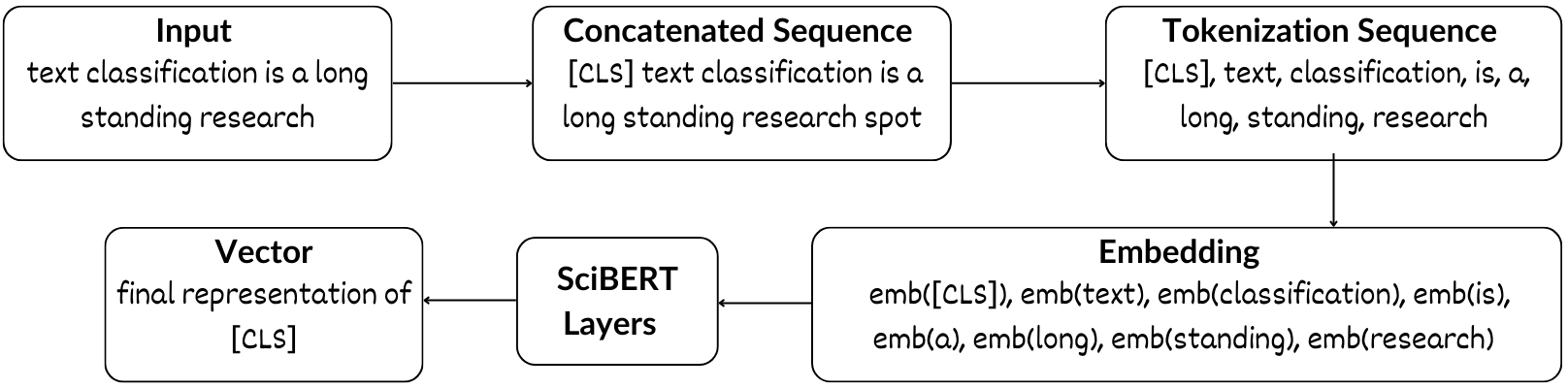}
\caption{Overview of SPECTER vectorisation} \label{Fig-7}
\end{figure}

Next up was to embed the chunks. As the embedding model, the pre-trained sentence transformer model SPECTER (allenai-specter) was used since it is trained to produce embedding for scientific documents~\cite{a27}. Unlike other transformers, SPECTER incorporates inter-document context into the transformer language models to learn document representations~\cite{a27}. To give a brief idea of the vectorisation process; SPECTER will encode the abstracts using the  SciBERT~\cite{a42} transformer and then take the final representation of [CLS] token as the final representation of the abstract~\cite{a27} as indicated in Fig.~\ref{Fig-7}. 

Finally, the embeddings were stored using the FAISS (Facebook AI Similarity Search) vector store, which is claimed to facilitate efficient similarity searches for natural language queries~\cite{a28}. It is worth noting that FAISS, implemented in C++, is an open-source similarity search space developed by META that comes with wrappers for Python~\cite{a29}. 

\subsection{Implementing the QA Pipeline}

\begin{figure}[ht]
\includegraphics[width=\textwidth]{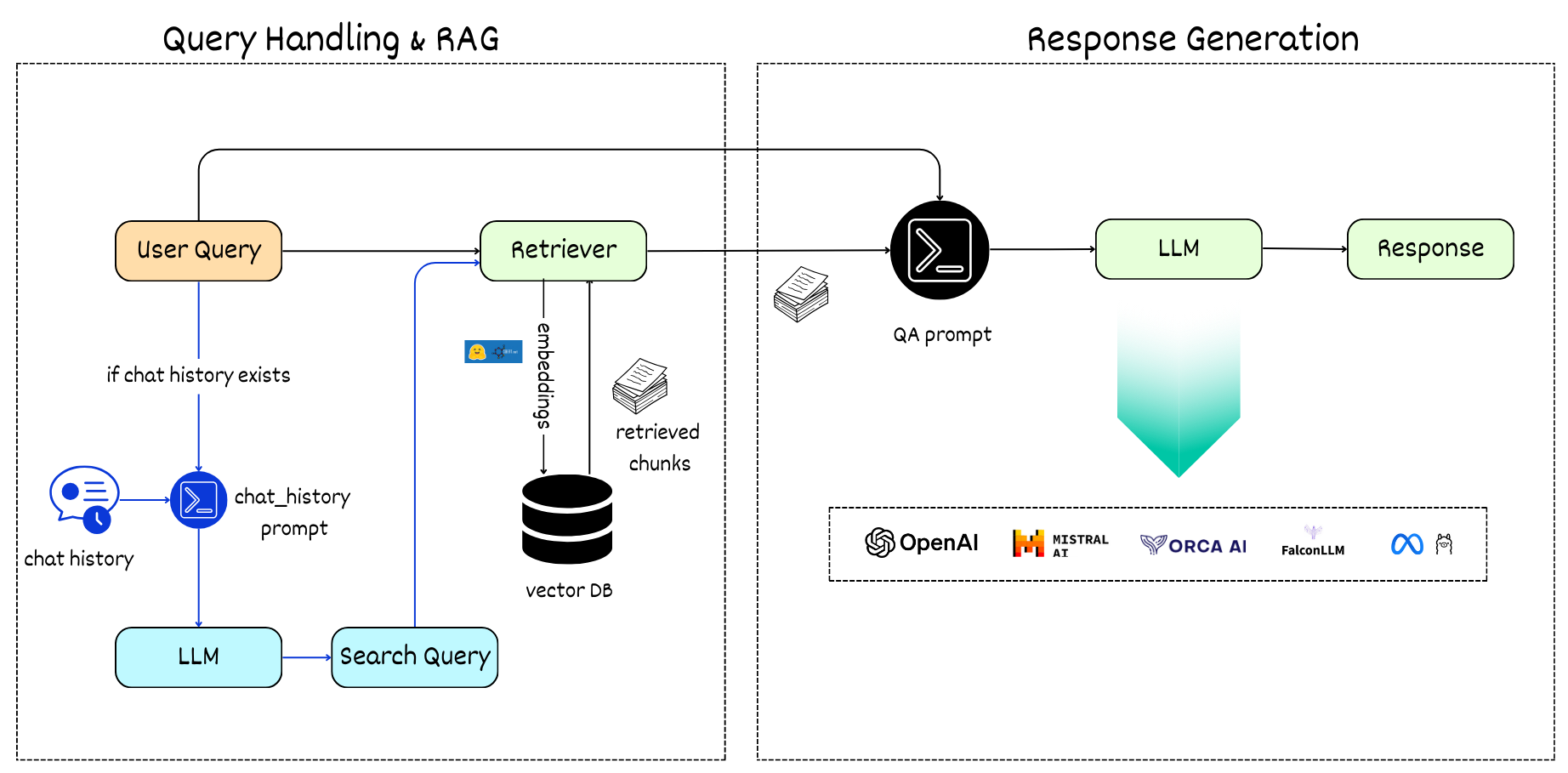}
\caption{Architecture of QA pipeline} \label{Fig-5}
\end{figure}

Upon building a vectorstore comprising abstracts from computer science literature, everything was set to design the QA pipeline. This is the most focused part of the study as it involved many brainstorming sessions, from deciding the LLMs to engineering the model instructions. The intention behind the QA pipeline was to get user queries as input, perform RAG on the vectorstore, and generate concise answers with the support of an LLM. Fig.~\ref{Fig-5} paints an overall picture of how the proposed QA pipeline works. 

Once a user inputs their query, the algorithm will check if a chat history exists for the current session. If chat history exists, then the question and the chat history will be passed to an LLM to get a refined search query, which is then introduced to the retriever. From the point of retrieving data to the response generation, the process will be similar, as explained below, irrespective of the existence of chat history. 

Regardless of the existence of a chat history, the user query will be sent to both the retriever and the QA prompt simultaneously. Once the retriever receives a user query, it will be embedded into a vector using the embedding model SPECTER. Then, the query vector will be mapped to the chunks in the vector store using a similarity search algorithm, where the threshold is set as 0.6 to retrieve only more similar chunks and k as 10 to return the top 10 most similar chunks. Next, the retrieved chunks and the user query formatted by the QA prompt are passed to an LLM to generate a concise response. 

Utilising a vector store along with an LLM facilitates semantic search. Semantics refer to the contextual meaning of a search query~\cite{a30}, and with a combination of embeddings, vector indexes and a similarity algorithm as proposed above, it is possible to carry out a semantic search~\cite{a31}. 

Since this study aims to compare the performance among multiple LLMs, it was decided to evaluate the QA capabilities of 5 of the most trending LLMs at the time. Among the LLMs tested were the leading pay-as-you-go LLM GPT-3.5 and the open-source LLMs: LLaMa 2-7b-chat, Mistral-7b-instruct, Flacon-7b-instruct and Orca-mini-v3-7b. It is worth noting that only quantised versions of open-source LLMs were called to the local environment since they improve the computational efficiency for a little trade-off in performance~\cite{a32}. 

\begin{figure}[ht]
\includegraphics[width=\textwidth]{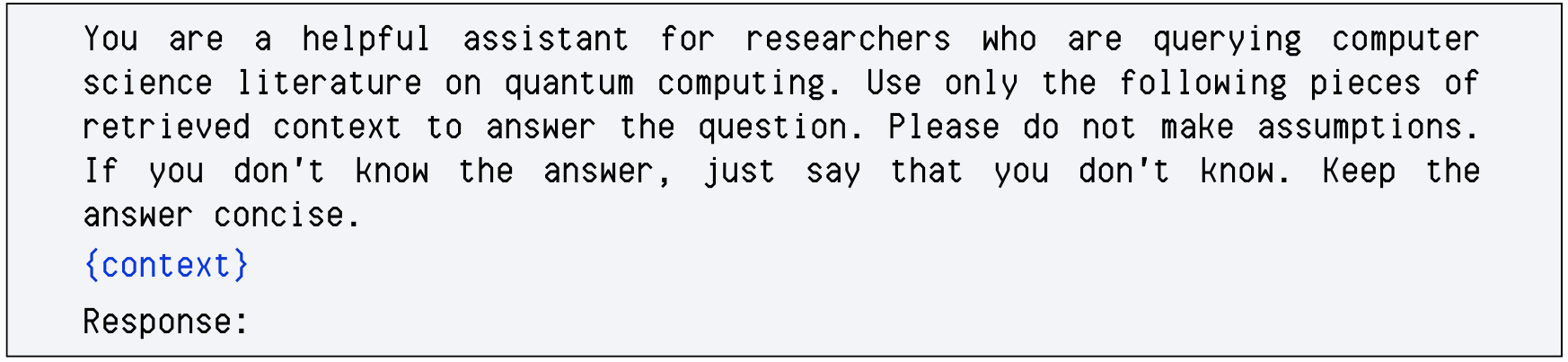}
\caption{Prompt designed for GPT-3.5 to query about quantum computing} \label{Fig-6}
\end{figure}

Regarding incorporating an LLM, there is another important aspect to point out: prompt engineering. As McKinsey \& Company~\cite{a33} has elaborated, to get a well-structured response using GenAI, it needs patience and iteration to provide clear instructions that lead to a precise answer. Prompt engineering is the concept of defining the ideal instruction for an LLM. Thus even in the underlying study, a special attention was paid on prompt engineering to design a prompt that generates answers for user queries using computer science literature.  Fig.~\ref{Fig-6} indicates the prompt passed to GPT-3.5 to generate long-form answers to queries on quantum computing. 

Fig.~\ref{Fig-5} shows two prompts being called at two different instances. To clarify, if a session already has a chat history, then the ‘chat\_history’ prompt will be passed with the chat history and the user query to generate an extended search query to be integrated to the retriever. Apart from that, the QA prompt is called regardless of the existence of a chat history to pass the user query and generate an answer from the retrieved content.  

The architectures offered by the LangChain community~\cite{a26} were used to design the above-discussed conversational RAG chain. In the proposed application, three chains will be maintained: one to structure the retrieved content (referred to as the format chain), another one to deal with conversation history (referred to as the history chain), and the other to perform RAG and generate responses (referred to as the QA chain). For the format chain, “create\_stuff\_documents\_chain” was used, which will format all the retrieved contents into a single prompt and pass it to an LLM~\cite{a26}. Since there is a concern about maintaining conversations (allowing a user to ask follow-up questions), for the history chain, LangChain’s “create\_history\_aware\_retriever” was used as it takes chat history into account and creates a search query to be passed into the retriever~\cite{a26}. Both the history chain and format chain are called within the QA chain built with “create\_retrieval\_chain”, which passes the user query to the retriever, fetches relevant chunks and then sends the user query along with retrieved chunks to the LLM for the response~\cite{a26}. 

With everything above set, it took just one function call to obtain the user’s input and generate a concise answer.

\subsection{Performance Evaluation}

Since one of the major focuses of the study is to evaluate the performances of multiple LLMs when employed with RAG, a custom dataset comprising 30 question-answer pairs was designed. These question-answer pairs were scripted by two human experts covering the latest concerns in the computer science field based on the same set of articles used to create the vector database. Further to that, the answers were crafted by human experts based on the content of the abstracts. The above mechanism of crafting QA pairs was inspired by the work conducted by L\'ala et al.~\cite{a7}, who proposed 50 QA pairs using biomedical literature that comes after September 2021 to check the functionality of their RAG-enabled LLM. The QA pairs proposed in this study span across the three research areas: quantum computing, large language models, and edge computing, with ten pairs for each area. The quantum computing  QA set was explicitly crafted with follow-up questions that required long-form answers. All the remaining QA pairs consisted only of Yes/No questions (binary questions). When developing the dataset, it was assumed that the questions are novel and, thus, have lower chances of being included in the training of base LLMs, requiring RAG to generate accurate answers~\cite{a7}. A few of the sample questions are indicated in Table~\ref{tab1}. 

\begin{table}[ht]
\centering
\caption{Sample questions for evaluation.}\label{tab1}
\begin{tabular*}{\textwidth}{@{\extracolsep{\fill}}ll}
\hline
\textbf{Question}                                                                                           & \textbf{Expected Answer Type} \\ \hline
\parbox[t]{0.67\textwidth}{01.Are there any studies to prove that quantum algorithms outperform classical algorithms in portfolio optimisation?} & Yes/No                        \\ 
02. What are the results of those studies?                                                                     & Follow-up question  \\
03. What is the goal of post-quantum cryptography?                                                              & Long answer                   \\
\hline
\end{tabular*}
\end{table}

With a custom QA dataset to test the performance of LLMs, two approaches were adopted to evaluate these questions. When generating answers for the binary questions, LLMs were instructed to generate either “yes”, “no” or “do not know”. Accuracy (\ref{eq:1}) and precision (\ref{eq:2}) were used to evaluate the LLMs’ performance in answering binary questions.

\begin{equation}
\text{Accuracy} = \frac{\text{No. of correct answers}}{\text{Total no. of questions}}
\label{eq:1}
\end{equation}

\begin{equation}
\text{Precision} = \frac{\text{No. of correct answers}}{\text{No. of confidently answered questions}}~\cite{a7}
\label{eq:2}
\end{equation}

Average cosine similarity was used to evaluate LLM performances in answering the questions that expect long-form answers. Cosine similarity (\ref{eq:3}) is a similarity measure that calculates the distance between two embeddings~\cite{a38} (for the underlying study, it will be the candidate answer and the generated answer). When the cosine similarity is one, it is stated that two vectors are identical, while 0 will imply the two vectors are opposed~\cite{a38}. Here, the average cosine similarity will be the summation of similarity scores averaged across the total number of questions. 

\begin{equation}
\text{Cosine Similarity}~\cite{a7} = \frac{A \cdot B}{\|A\|\,\|B\|}
\label{eq:3}
\end{equation}

Apart from that, long-answer questions were evaluated by a human expert and the AI chatbot Gemini, which was developed by Google. Both the human and AI bot were asked to compare the generated answer with an answer candidate and rate the usefulness of the generated answer. The ratings included three levels: poor, average, and excellent. Poor points to the answers that do not meet the minimum expectations of the answer candidate, while excellent means answers that satisfy all the expectations of the answer candidate.

All the above-mentioned evaluation metrics will consider the usefulness of an answer. Thus, keeping the usefulness of the answers aside for a moment, towards the end of the study it was decided to focus on latency (time taken to generate an answer) and the cost incurred in generating the answer as some side parameters supporting further comparison between the LLMs.  

%Results 

\section{Results and Discussion}
\label{sec:Discussion}

This section will comprehensively evaluate the performance of the LLMs in answering binary questions and long-answer questions. All the trials were carried out on a MacBook Pro with an M2 processor and 16GB of RAM. Moreover, same values were set to the parameters of all the LLMs to ensure comparability. The LLM temperature was set to 0.01 since the underlying task is fact-based QA, which does not encourage creative answers but rather more concise answers, while the ‘max\_tokens’, which defines the number of tokens the model will process, was set to 2000. 

With the above configurations, the results obtained under each LLM for binary questions and long-answer questions are indicated in Table~\ref{tab2} and Table~\ref{tab3}. Table~\ref{tab2} indicates the accuracy and precision calculated after evaluating the results generated for 21 binary questions. Accordingly, it is observable that GPT 3.5 combined with RAG has reported the highest accuracy and precision. This will reconfirm that GPT is the all-time best LLM for most natural language tasks. The reason behind its remarkable performance is that GPT-3.5 is trained on 175 billion parameters~\cite{a34}, which is more significant when compared to most of the existing LLMs.\footnote{GPT-4 is trained on even more parameters thus, it will definitely surpass GPT 3.5. Since, in this study, the focus is open-source LLMs, the inclusion of GPT 3.5 is to give an overall idea of how the results would vary if we go for a larger LLM.} However, as indicated in Table~\ref{tab4}, GPT 3.5 has a cost. To generate answers for 30 questions with a maximum of 2000 tokens in each run, it costs \$0.000643. 

Moving to the open-source LLMs and their performance in answering the binary questions, the Mistral-7b-instruct model combined with RAG surpassed the rest in accuracy and precision. Compared to GPT-3.5, Mistral-7b-instruct, with 7 billion training parameters, has reported an accuracy of 0.857, whereas GPT 3.5+RAG recorded its accuracy as 0.9048. This is a promising result for an open-source LLM and highlights that these LLMs can be relied upon to address most of the business needs. These LLMs are entirely free; thus, no financial cost was involved in generating answers. However, in Table~\ref{tab4}, it is clear that GPT-3.5 has taken only 1.7413 seconds on average to create an answer, whereas Mistral-7b-instruct took 105.9543 seconds on average to generate an answer. Thus, the main trade-off is between latency and the cost involved, while a small trade-off concerns the answer’s usefulness. However, it is worth noting that an enhancement to local hardware resources will help to reduce the latency of open-source LLMs (ex: accelerating CPU to GPU).

Following the Mistral-7b-instruct, the Orca-mini-v3-7b model also shows promising results for binary questions with a recorded accuracy of 0.8095. When it comes to precision (the total number of correct answers over confidently answered questions), both LLaMa-2-7b-chat and Falcon-7b-instruct show relatively lower performances. This implies; though these models are compiled with RAG, there are still factually incorrect yet confidently answered questions. This is called hallucination, where LLMs generate plausible-sounding but unfaithful or nonsensical information~\cite{a39}. 

\begin{table}[ht]
\centering
\caption{Performance metrics for binary questions.}\label{tab2}

\begin{tabular*}{\textwidth}{@{\extracolsep{\fill}}lll}

\hline

\textbf{LLM}           & \textbf{Accuracy} & \textbf{Precision} \\\hline
GPT 3.5 + RAG & 0.9048   & 0.9048    \\
Orca + RAG    & 0.8095   & 0.8095    \\
LLaMa 2 + RAG & 0.6190   & 0.6842    \\
Falcon + RAG  & 0.619    & 0.6842    \\
Mistral + RAG & 0.8571   & 0.8571    \\
Chat-GPT      & 0.4761   & 0.625    \\\hline
\end{tabular*}
\end{table}

To evaluate long answers, this study utilised a cosine similarity score, an AI-generated rank (Gemini), and a human-expert evaluated rank across. Cosine similarity values indicated in Table~\ref{tab3} are relatively low, signalling that the generated answers are different from the answer candidates in terms of similarity. This can be due to the rich vocabulary of LLMs that lets them generate answers more creatively. As per the results indicated in Table~\ref{tab3} GPT 3.5+RAG, has the highest cosine similarity score of 0.4479, and among the open-source LLMs, Mistral-7b-instruct has the highest score of 0.2754.  

Since the cosine similarity scores are lower, it alone can not evaluate the performance of LLMs in long-form question answering. Therefore an AI-generated rank and a human-evaluated rank were combined to support the overall evaluation. For the human evaluation, an expert in the quantum computing field was employed. The expert was provided with the question,  the answer candidate and the relevant abstract and was asked to evaluate the generated answer with the help of the provided content and her expertise. To get Gemini rankings, a clearly crafted instruction was given to compare the answer candidate and the generated answer and rank the generated answer based on its quality.

As indicated in Table~\ref{tab3}, Mistral-7b-instruct has performed well compared to other open-source LLMs, with 6 out of 9 answers being ranked excellent by both Gemini and the human expert. From the open-source LLMs being compared, the LLM with the weakest performance in terms of similarity score was Falcon-7b-instruct and was the same under both Gemini rankings and human rankings. This indirectly implies that language tools like Gemini are optimised considerably to replace the human capacity to understand the context. All in all, among the LLMs tested, GPT 3.5 combined with RAG, has a better performance in long-form question answering (Table~\ref{tab3}).

\begin{table}[ht]
\centering
\caption{Performance metrics for long-answer questions.}\label{tab3}
\begin{tabular*}{\textwidth}{@{\extracolsep{\fill}}llcccccc}

\hline
\multicolumn{1}{c}{\multirow{2}{*}{\textbf{LLM}}} & \multirow{2}{*}{\textbf{\begin{tabular}[c]{@{}c@{}}Cosine \\ Similarity\end{tabular}}} & \multicolumn{3}{c}{\textbf{AI Rank (Gemini)}}                                                 & \multicolumn{3}{c}{\textbf{Human Rank}}                                                                           \\ \cline{3-8} 
\multicolumn{1}{c}{}                              &                                                                                        & \textbf{Poor} & \multicolumn{1}{l}{\textbf{Average}} & \multicolumn{1}{l}{\textbf{Excellent}} & \multicolumn{1}{l}{\textbf{Poor}} & \multicolumn{1}{l}{\textbf{Average}} & \multicolumn{1}{l}{\textbf{Excellent}} \\ \hline
GPT 3.5 + RAG                                     & 0.2291                                                                                 & 3             & 5                                    & 12                                      & 8                                 & 3                                    & 9                                      \\
Orca + RAG                                        & 0.1806                                                                                  & 7             & 6                                    & 7                                      & 11                                & 3                                    & 6                                      \\
LLaMa 2 + RAG                                     & 0.2263                                                                                  & 4             & 3                                    & 13                                      & 7                                 & 6                                    & 7                                      \\
Falcon + RAG                                      & 0.2326                                                                                 & 4             & 9                                    & 7                                      & 11                                 & 3                                    & 6                                      \\
Mistral + RAG                                     & 0.2339                                                                                 & 3             & 7                                    & 10                                      & 5                                & 5                                    & 10                                      \\
ChatGPT                                           & 0.0977                                                                                 & 7             & 5                                    & 8                                      & 12                                 & 4                                    & 4                                      \\ \hline
\end{tabular*}
\end{table}

There is an important fact to notice about the evaluation of long-form answers. Since the Gemini-AI does not have the privilege to access recent knowledge and the fact that it has informed knowledge~\cite{a43}, a slight change to the instructions resulted in contradictory ranks questioning the biasness in Gemini generated rankings. Moreover, if the same question was repeated multiple times, in some instances the AI-generated rank changed due to the associated complexity and variability of responses. Further to that, at present the study employs only a single human to get the expert opinion. Thus her understanding of quantum computing and the point-of-view might include a slight bias in the above rankings. However, a comparison between the expert-generated rankings and the AI-generated ranking shows that the rankings are almost line-in-line leaving a little or zero bias in the evaluation scores. As the study expands, it is proposed to utilise a pool of human expertise and average out their rankings to reduce the bias in human score. 

Though the main focus was to compare the performances of  LLMs when combined with RAG, it was decided to test how the answers would look like without RAG. Therefore, ChatGPT, which is OpenAI’s most demanding chatbot built on GPT-3.5, was asked the same questions. The same prompt template and a new instance of a chat were used to generate answers for each question. Under this scenario, the results obtained for accuracy and precision were the worst compared to all the other RAG-enabled LLMs. Though ChatGPT is one of the best AI chatbots, it lacks the skill to answer questions on recent content, underscoring that off-the-shelf models are not carrying out RAG if not customised. 

Keeping GPT-3.5 aside, comparing the latency among the open-source LLMs shows that the Orca-mini-v3-7b model surpasses the rest with an average latency of 99.2167 seconds (Table~\ref{tab4}). Out of all the open-source LLMs tested in the study, Meta’s LLaMA-2-7b-chat has the highest average latency in generating answers (Table~\ref{tab4}). 

\begin{table}[ht]
\centering
\caption{Average latency and average cost across LLMs.}\label{tab4}

\begin{tabular*}{\textwidth}{@{\extracolsep{\fill}}lll}
\hline
\textbf{LLM}                                  & \textbf{Average Latency} & \textbf{Average Cost}     \\\hline
GPT-3.5 + RAG                         & 1.7413          & 0.000643         \\
Orca + RAG                            & 99.2167         & No cost involved \\
LLaMa2 + RAG                         & 107.4748        & No cost involved \\
Falcon + RAG                          & 107.31642       & No cost involved \\
Mistral + RAG & 105.9543        & No cost involved \\\hline
\end{tabular*}
\end{table}

It is worth noting that some LLMs required prompts to follow a specific format to generate better answers (ex: Mistral-7b-instruct and falcon-7b-instruct). In contrast, others (ex: GPT-3.5) understand natural language easily without needing a specifically structured prompt. In the underlying study, when given a prompt to Mistral-7b-instruct without following its template \footnote{<s>[INST] instructions [/INST] answer </s>[INST] Follow-up instructions [/INST]}, it failed to generate answers as expected. However, structuring the prompt to fit the template made it possible to improve the answer, as indicated in Table~\ref{tab5}.

\begin{table}[ht]
\centering
\caption{Comparison between structured \& unstructured prompts for Mistral.}\label{tab5}

\begin{tabular*}{\textwidth}{@{\extracolsep{\fill}}ll}
\hline
\textbf{Question}                                   & \parbox[t]{0.85\textwidth}{Are there any studies proposing Lightweight Unified Collaborated Relinquish Edge Gateway Architecture with isochronal patching?}                                                                                                                                                                                                                                                                                                                                                                                                                                                                                                                                                  \\\hline
\parbox[t]{0.15\textwidth}{\textbf{Answer (not following the prompt template)}} & \parbox[t]{0.85\textwidth}{Based on the context provided, I cannot directly answer that question with a simple yes or no. However, I can tell you that the text you provided discusses various aspects of edge computing, including architectures, advantages, enabling technologies, and standardization. However, it does not specifically mention or propose a "Lightweight Unified Collaborated Relinquish Edge Gateway Architecture with isochronal patching" as such. If you have more specific information or context, please provide it and I will do my best to help you with your question.}                                                                                                             \\\hline
\parbox[t]{0.15\textwidth}{\textbf{Answer (following the prompt template)}}     & \parbox[t]{0.85\textwidth}{Based on the provided context, there is indeed a study that proposes a lightweight unified collaborated relinquish (LUCR) edge gateway architecture with isochronal patching. This study aims to manage the dynamic updation process and increase the processing capacity and isochronal patching is utilized to prevent backdoor functionality problems and also manage the considerable amount of hardware and software resources with its lightweight architecture. However, it's important to note that this is just one study, and there may be other approaches and architectures that also address the challenges and requirements of managing and processing data at the edge.}\\\hline
\end{tabular*}
\end{table}

From the above discussion, it is obvious that incorporating RAG on top of an LLM allows a user to query publications in the recent past. Thus, unlike an off-the-shelf LLM, a RAG-based QA system powered by an LLM for computer science literature has the special ability to answer user queries in the computer science domain with the most recent and valid content while ensuring the novelty of the answer. 

Further to that, looking into Mistral’s success and Falcon’s failure as open-source LLMs, the difference in attention mechanisms can be pointed out. The attention mechanism allows a model to focus on the most relevant parts of the input when producing the output~\cite{a44}. Llama2 utilises the vanilla multi-head attention (MHA) which runs through the attention mechanism several times in parallel requiring  a high memory bandwidth~\cite{a44}. Falcon uses a multi-query attention (MQA) mechanism~\cite{a46} which is more efficient in memory load at the cost of output quality. On the other hand, Mistral uses a grouped query attention (GQA) which results in higher throughput by allowing for higher batch sizes. Moreover, the sliding window attention in Mistral is designed to handle longer sequences more effectively at a reduced computational cost~\cite{a45}. As GQA stand in the balance between MQA and MHA, it is fair to expect a significant performance improvement from the Mistral model in terms of both answer quality and latency.

% ---- Conclusion -----
\section{Conclusion}
\label{sec:Conclusion}
Question-answering systems are used widely worldwide in e-learning platforms, social media applications, customer support platforms, and more. It has expanded its horizons with the introduction of LLMs owing to their creativity and the ability to understand natural language queries. One area that could benefit from QA is scientific literature. As it is time-consuming for the researchers to go through existing literature one by one, a system that allows them to query the published work will be of immense advantage. However, simply incorporating an off-the-shell LLM won’t generate results as expected due to the limitations in training.  This study has integrated RAG on LLMs to assess their performance in question answering on computer science literature.

LLM-based chatbots still struggle with hallucination because of the limitations they have in accessing recent knowledge bases~\cite{a35}, leading chatbots to provide false or misleading answers. For example, when it comes to ChatGPT, it gives incorrect information to questions that were raised in the recent past because it has been trained on data till January 2022. Further, it is not proficient in providing references to the most recent publications, which doubts the validity of the answers generated~\cite{a36}. By introducing RAG to LLMS, these limitations with outdated training data can be eliminated. The study shows that GPT-3.5 integrated with RAG delivers better answers than standalone ChatGPT. As recently published (from 2023 to 2024) scholarly articles on computer science were used to implement RAG, LLMs could answer the questions connected with the recent past. Also, they can prove the answers by referring to literature.  To summarise, integrating RAG enables LLMs to access recent knowledge sources to generate answers. With the rich textual context that LLMs hold, it can provide more creative, non-hallucinated, logically correct answers.

With the emergence of LLMs, prompt engineering has become a trending topic. It underlies the concept of designing, refining, and implementing prompts for LLMs to deliver optimal outputs~\cite{a37}. Therefore, it is fair to state that a prompt affects the quality of outputs generated by LLMs, and it is essential to provide clear and precise instructions for LLMs to perform the tasks effectively. Even in the underlying study, it was required to develop multiple prompts before choosing a single prompt to instruct the LLMs. It appears as another area that needs a thorough evaluation. Thus, the prompt used for the study might not be optimal, but the results prove that it certainly is a good enough option. 

As the study revolves mainly around open-source LLMs, the users do not need to worry about the cost. However, the most significant limitation in most of these open-source LLMs is the latency that comes with them. This can be easily avoided by going for a subscription-based LLM that enables LLM invocation through infrastructure provided by the hosting company. On the other hand, one can accelerate their local infrastructure when calling an open-source LLM to reduce its latency. However, the cost associated with each instance should be considered depending on the use case. For example, in this study, an LLM is combined with RAG to allow individuals interested in computer science to query published literature easily. Suppose the application values accuracy above cost, then, instead of accelerating infrastructure to host an open-source LLM, it is advised to go for a model in the GPT family. If the concern is more towards data privacy, a GPU-accelerated infrastructure will be the ideal option, as it will let you run an open-source LLM efficiently in your local environment. However, if the intention is personal use, incorporating a Mistral-7b-instruct-like model with RAG under the available infrastructure will generate quality answers within a reasonable time.

Overall, this study provides a comprehensive evaluation of the performance of some of the most trending LLMs in question-answering on computer science literature. Among the models compared, the study concludes that GPT 3.5 is the best-performing LLM overall, and Mistral-7b-chat is the best in open-source LLMs.  However, since the field of LLMs keeps evolving, and different models emerge with diverse parameters and capabilities, it is essential to keep up with the field and check for the usability of emerging LLMs before deciding on an LLM to be integrated into your application.

Lastly, the underlying study allows future researchers to extend it further by incorporating prompt engineering, trying out different mechanisms to retrieve data in contrast to vector stores, utilising full-text research articles rather than limiting to abstracts and even expanding its applicability to other domains.

% ---- Bibliography ----

\bibliographystyle{splncs04}
\bibliography{references}
\end{document}